\documentclass[letterpaper, 10 pt, conference]{ieeeconf}

\IEEEoverridecommandlockouts

\usepackage{microtype}
\usepackage{graphicx}
\usepackage{subcaption}
\usepackage{booktabs}
\usepackage{times}
\usepackage{epsfig}
\usepackage{amsmath}
\usepackage{amssymb}
\usepackage{etoolbox}
\usepackage{multirow}
\usepackage{listings}
\usepackage{xcolor}
\usepackage{colortbl}
\usepackage{xspace}
\usepackage{mathtools}
\usepackage[colorlinks=true,linkcolor=black,citecolor=black,urlcolor=black]{hyperref}

\definecolor{ourslightblue}{RGB}{232,242,255}

\newcommand{\method}{PackLab\xspace}

\title{\LARGE \bf
\method: A Comprehensive Framework for \\ Developing, Training, and Evaluating MLLMs in Robotic Bin Packing
}

\author{Donghao Zhou$^{1,*}$, Jia-Hui Pan$^{1,*}$, Fan Zhang$^{1}$, Xingyuan Bu$^{2}$, Shilong Li$^{2}$, Xiaojie Gao$^{1}$,\\
Yun-Hui Liu$^{3}$, Chi-Wing Fu$^{1,\dagger}$, and Pheng-Ann Heng$^{1,\dagger}$%
\thanks{This study was supported by the InnoHK initiative of the Innovation and Technology Commission of the Hong Kong Special Administrative Region Government via the Hong Kong Centre for Logistics Robotics.}
\thanks{$^{*}$Equal contribution. $^{\dagger}$Corresponding authors.}
\thanks{$^{1}$Donghao Zhou, Jia-Hui Pan, Fan Zhang, Xiaojie Gao, Chi-Wing Fu, and Pheng-Ann Heng are with the Department of Computer Science and Engineering, The Chinese University of Hong Kong.}
\thanks{$^{2}$Xingyuan Bu and Shilong Li are with M-A-P.}
\thanks{$^{3}$Yun-Hui Liu is with the Department of Mechanical and Automation Engineering, The Chinese University of Hong Kong.}
}

\begin{document}

\bstctlcite{BSTcontrol}
\maketitle
\thispagestyle{empty}
\pagestyle{empty}

\begin{abstract}
Robotic bin packing requires long-horizon sequential decision-making, as each object placement affects the available space for subsequent packing.
Existing methods primarily rely on hand-crafted geometric heuristics that optimize predefined objectives or reinforcement learning policies learned through trial and error over predefined training configurations.
Despite recent advances in multimodal large language models (MLLMs) for this task, their potential for closed-loop sequential decisions across heterogeneous packing configurations remains underexplored.
To address this gap, we introduce \textbf{PackLab}, a comprehensive framework for developing, training, and evaluating MLLMs for closed-loop robotic bin packing.
\textit{PackLab-Suite} provides a physics-based simulation platform for scalable generation of diverse training packing trajectories and evaluation of their physical outcomes.
\textit{PackLab-VLM} is a packing-specialized MLLM that understands the evolving object and container states to jointly select objects and predict placements in a closed-loop manner.
\textit{PackLab-Bench} provides standardized packing scenarios at multiple difficulty levels for systematic evaluation.
Extensive experiments demonstrate that, on average, \textit{PackLab-VLM} outperforms conventional packing heuristics, traditional reinforcement learning methods, and general-purpose MLLMs across object sets and container configurations, highlighting the potential of MLLMs for long-horizon robotic packing.
The code, model, dataset, and benchmark are available at \href{https://github.com/Correr-Zhou/PackLab}{\textcolor{purple}{\nolinkurl{https://github.com/Correr-Zhou/PackLab}}}.
\end{abstract}

\section{Introduction}

Robotic bin packing is a classical optimization problem that maximizes space utilization by arranging objects into a constrained container, with broad applications in logistics, warehousing, and robotic manipulation.
It is not only a geometric optimization problem, but also a sequential decision-making problem in which each object placement changes the available space for all subsequent objects.
Effective packing therefore requires understanding the long-term consequences of individual placement decisions.

Existing robotic packing methods primarily rely on packing heuristics~\cite{karabulut2004hybrid, ramos2016container, wang2019stable, wang2021dense, pan2023sdf} or reinforcement learning (RL)~\cite{huang2022planning,zhao2023learning, hu2020tap, zhao2021online, yang2023heuristics}.
Heuristic methods score candidate placements using hand-crafted geometric objectives, whereas RL methods learn sequential policies through trial and error, typically under predefined object and container distributions.
Learning a unified policy that handles heterogeneous packing configurations, however, remains challenging.
Recent advances in multimodal large language models (MLLMs) offer a promising foundation for learning a unified packing policy across heterogeneous configurations, as their multimodal representations and flexible sequence modeling enable joint conditioning on variable object sets, container geometries, and interaction histories.
Nevertheless, existing studies~\cite{pan2025opa,blei2025ipack, blei2025llm} primarily use MLLMs to infer object properties or semantic constraints, while relying on conventional planners to determine packing actions.
This raises a fundamental question: \textit{Can an MLLM directly perform closed-loop object selection and placement across diverse object sets and container configurations?}

\begin{figure}[t]
    \centering
    \includegraphics[width=\linewidth]{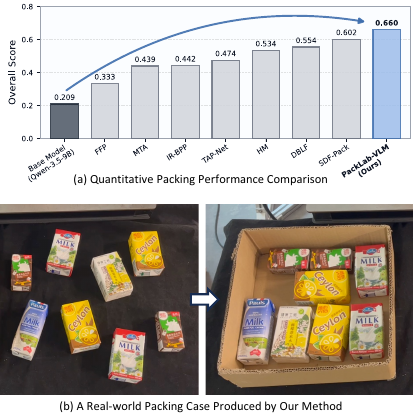}
    \caption{(a) \textit{PackLab-VLM} outperforms the baseline and existing methods. (b) A real-world packing result produced by our method, with 1.00 Success Ratio and 0.72 Compactness.}
    \label{fig:teaser}
\end{figure}

Answering this question requires developing and evaluating MLLM-based packing policies across diverse object sets, container configurations, and task complexities.
However, existing work lacks an integrated framework for physics-grounded simulation, scalable training data generation, packing-specific training, and standardized evaluation.
To address this gap, we introduce \textbf{PackLab}, a comprehensive framework for developing, training, and evaluating MLLMs in robotic bin packing.
In Figure~\ref{fig:teaser}, we illustrate PackLab's superior packing performance compared with existing methods in terms of the Overall Score, together with a real-world packing example to further show its effectiveness.

PackLab comprises three core components that respectively support physics-grounded simulation and scalable data generation, closed-loop packing-specific training, and standardized evaluation.
Specifically, \textit{PackLab-Suite} is a physics-grounded platform for packing simulation with a layout-first data generation engine to produce diverse, physically valid training packing trajectories, curated as the PackData-20K dataset.
\textit{PackLab-VLM} is fine-tuned on PackData-20K to integrate container heightmaps, candidate-object attributes, and observation-action histories, jointly predicting object and placement selection at each step.
Lastly, \textit{PackLab-Bench} structures test cases into three difficulty levels progressing from easy to hard through larger buffers, richer object sets, and enlarged container dimensions.
It provides standardized evaluation metrics of Success Ratio, Compactness, and their product as the Overall Score.
On average, \textit{PackLab-VLM} outperforms packing heuristics, RL methods, and the general-purpose MLLM baseline across heterogeneous packing settings.
Ablation studies validate the effectiveness of each design, while physical experiments validate its applicability to real-world robotic bin packing.

The main contributions of this work are as follows:
\begin{itemize}
    \item We introduce \textbf{PackLab}, a comprehensive framework
    for developing, training, and evaluating MLLM-based robotic
    bin packing policies.
    \item Three core components are developed within the proposed framework: \textit{PackLab-Suite} for physics-grounded simulation and scalable training trajectory generation, \textit{PackLab-VLM} for closed-loop packing decisions, and \textit{PackLab-Bench} for standardized evaluation.
    \item We conduct extensive evaluations, demonstrating \textit{PackLab-VLM}'s effectiveness across heterogeneous packing settings and its real-world feasibility.
\end{itemize}

\section{Related Work}

Existing robotic bin packing methods adopt different paradigms for packing optimization.
Early approaches optimize entire packing sequences offline using genetic algorithms~\cite{ramos2016container,karabulut2004hybrid,kang2012hybrid,gonccalves2013biased}, tabu search~\cite{lodi2002heuristic}, simulated annealing~\cite{liu2015hape3d}, or integer linear programming~\cite{lamas2022voxel,jiang2012learning}.
However, these approaches incur substantial computational costs and may require replanning when executed placements deviate from planned ones.

To enable online planning, packing heuristics evaluate candidate placements using hand-crafted geometric objectives~\cite{karabulut2004hybrid,wang2010two,wang2019stable,wang2021dense,pan2023sdf}.
Existing packing heuristics evaluate candidate placements using different criteria.
Deepest-Bottom-Left-Fill (DBLF) prioritizes placements that are both deep and low~\cite{karabulut2004hybrid}.
Maximum-Touching-Area (MTA) maximizes the contact area between the placed item and its surrounding objects and container walls~\cite{wang2010two}.
Heightmap-Minimization (HM) selects placements that minimize the increase in container heightmap~\cite{wang2019stable,wang2021dense}.
SDF-Pack chooses placements with minimum signed distance field values~\cite{pan2023sdf}.
These methods efficiently select placements based on the current container state in an online manner.
However, their local objectives primarily evaluate individual placements without explicitly accounting for their effects on the remaining space and feasibility of subsequent placements.

To optimize long-term packing decisions, reinforcement learning (RL) methods formulate packing as a sequential decision-making problem and learn policies that maximize cumulative rewards~\cite{hu2020tap, duan2018multi, zhang2021attend2pack, huang2022planning, zhao2023learning}.
Unlike packing heuristics that evaluate only the immediate quality of a placement, these methods can account for how each action changes the remaining free space and influences subsequent decisions.
Representative approaches employ deep Q-networks~\cite{verma2020generalized}, hierarchical or dueling architectures~\cite{hu2020tap,zhao2023learning}, and specialized objectives that encourage stable placements~\cite{huang2022planning,zhao2023learning} or incorporate heuristic guidance to improve exploration and solution quality~\cite{yang2023heuristics}.
However, policy learning requires extensive trial-and-error over predefined object and container settings, which may limit generalization to unseen settings.

Recent advances in multimodal large language models (MLLMs) have motivated their application to robotic bin packing, leveraging their capabilities in object understanding.
Existing studies~\cite{blei2025llm,blei2025ipack,pan2025opa} primarily employ MLLMs to infer object properties, physical relationships, or packing constraints, which are subsequently incorporated into conventional planners for object selection and placement.
Sim et al.~\cite{sim2025beyond} employ a large language model to generate packing heuristics, but report limited generalization across problem instances.
PackingGPT~\cite{you2026packinggpt} directly predicts sequential object placements, but does not perform closed-loop decision-making based on evolving packing states.

In contrast to explicit optimization, hand-crafted heuristics, and trial-and-error policy learning, we aim to train an MLLM to jointly select objects and predict placements across diverse object and container size configurations.
Compared with prior MLLM-based approaches, PackLab performs long-horizon, closed-loop robotic packing by understanding the dynamically evolving container and object states.

\section{Methodology}
\label{sec:method}

\begin{figure*}[t]
    \centering
    \includegraphics[width=\linewidth]{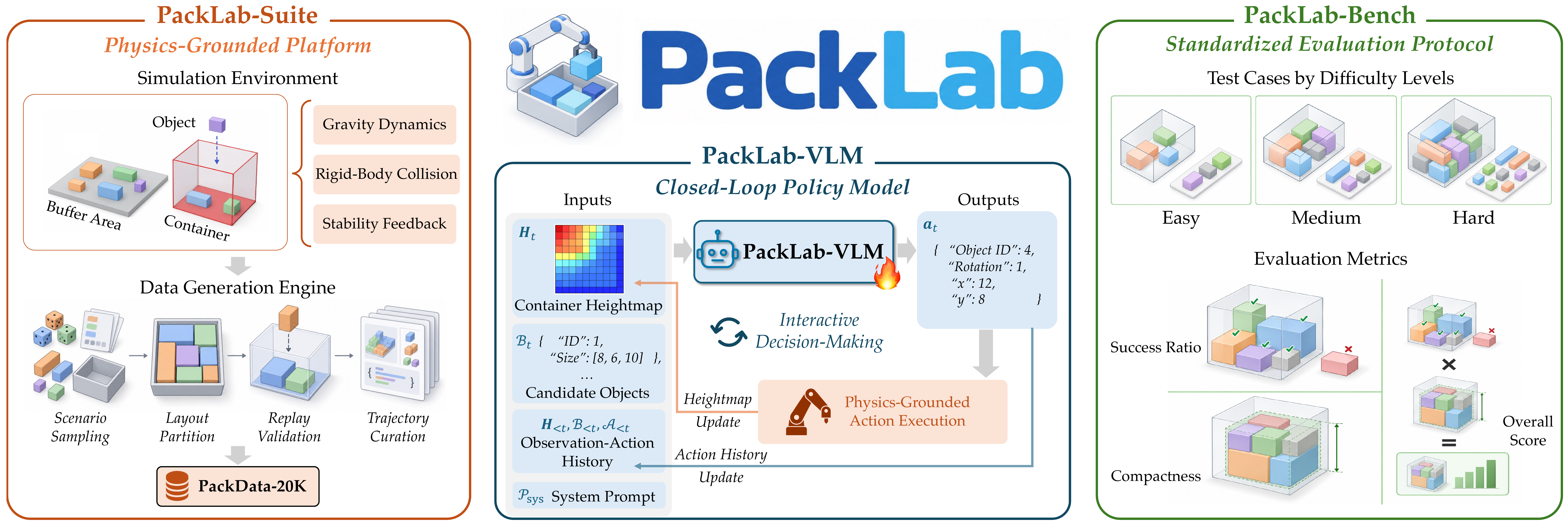}
    \caption{Overview of PackLab.
    \textit{Left:} \textit{PackLab-Suite} simulates physical interactions to generate, validate, and curate diverse packing trajectories to support policy training and form the PackData-20K dataset.
    \textit{Center:} \textit{PackLab-VLM} predicts packing actions in a closed-loop manner based on the container heightmap, candidate object attributes, observation-action history, and system prompt.
    \textit{Right:} \textit{PackLab-Bench} organizes test cases by difficulty levels and evaluates Success Ratio, Compactness, and Overall Score.}
    \label{fig:pipeline}
\end{figure*}

\subsection{Problem Formulation}

Robotic bin packing is formulated as a closed-loop sequential decision process over $T$ packing steps.
At step $t$, the scene state $\mathcal{S}_t=(\mathbf{H}_t,\mathcal{B}_t)$ comprises the current container occupancy, encoded as a top-down heightmap $\mathbf{H}_t$, and a candidate-object buffer $\mathcal{B}_t$ with each object described by its identity and 3D dimensions.
A packing policy predicts the packing action:
\begin{equation}
    \mathbf{a}_t = (b_t, o_t, x_t, y_t),
\end{equation}
where $b_t \in \mathcal{B}_t$ is the selected object, $o_t$ is its horizontal orientation, and its placement location $(x_t,y_t)$, which is defined by the minimal horizontal location of its bounding box at the target pose.

The vertical placement coordinate $z_t$ is determined by the lowest physically reachable position under gravity, which is written as
\begin{equation}
    z_t
    =
    \max_{\substack{
        0\leq i<l_{{b}_t}^{o_t}\\
        0\leq j<w_{{b}_t}^{o_t}
    }}
    \mathbf{H}_t[x_t+i,y_t+j],
    \label{eq:vertical_coordinate}
\end{equation}
where $l_{b_t}^{o_t}$ and $w_{b_t}^{o_t}$ denote the length and width of object $b_t$ at orientation $o_t$.
$\mathbf{H}_t$ denotes the top-down container heightmap.
After executing $\mathbf{a}_t$, the environment returns an updated scene state $\mathcal{S}_{t+1}$, which informs the next decision.
The packing process continues until all objects have been processed.
The goal is to maximize the successfully packed object volume and the compactness of the resulting object arrangement.

\subsection{Framework Overview}

PackLab is a self-contained robotic bin packing framework comprising \textit{PackLab-Suite}, a physics-grounded platform; \textit{PackLab-VLM}, a multimodal closed-loop policy; and \textit{PackLab-Bench}, a standardized evaluation protocol.
Figure~\ref{fig:pipeline} presents an overview of the framework.

\textit{PackLab-Suite} models gravity dynamics, rigid-body collisions, and stability feedback to facilitate the state transition from $\mathcal{S}_t$ to $\mathcal{S}_{t+1}$.
Training packing trajectories (\textit{i.e.}, the PackData-20K dataset) are constructed based on its data generation engine and are subsequently used to train the closed-loop policy model \textit{PackLab-VLM}.
At each packing step $t$, \textit{PackLab-VLM} predicts an action $\mathbf{a}_t$, specifying the selected object, orientation, and planar placement coordinates, based on the scene state $\mathcal{S}_t$ including the container heightmap and candidate object attributes, as well as the observation--action history.
After physics-grounded execution, the updated heightmap and action history are fed back to support closed-loop decision-making.
\textit{PackLab-Bench} organizes test cases into three difficulty levels, easy, medium, and hard, and evaluates the Success Ratio, Compactness, and Overall Score.

\subsection{\textit{PackLab-Suite}: Physics-Grounded Platform}

\textit{PackLab-Suite} is a physics-grounded platform that performs simulation with gravity dynamics, rigid-body collision, and stability feedback.
Its data generation engine constructs PackData-20K by partitioning container space into candidate layouts and inversely reconstructing the corresponding packing trajectories to facilitate subsequent policy training.

\vspace{+5pt}
\noindent \textbf{Simulation Environment.} \
After each placement $\mathbf{a}_t$, \textit{PackLab-Suite} simulates gravity-driven dynamics with the gravitational acceleration set to $-9.8\,\mathrm{m/s^2}$ along the vertical axis, allowing the newly placed object to settle within the container.
The objects, buffer surface, and container boundaries are represented using box-shaped collision geometries, enabling the simulator to resolve object--object, object--ground, and object--wall contacts.
The stability-feedback module advances the simulation while monitoring the placed object's linear and angular velocities; once both fall below predefined thresholds, the object is deemed stationary, and the resulting post-placement state is recorded to facilitate closed-loop planning.

\vspace{+5pt}
\noindent \textbf{Data Generation Engine.} \
\textit{PackLab-Suite} also incorporates a data generation engine that adopts a layout-first data generation pipeline that avoids exhaustive forward sampling over object--placement combinations.
The data generation procedure comprises four stages: scenario sampling, layout partition, replay validation, and training trajectory curation.

First, the container size and the number of packing objects are sampled to instantiate each packing scenario.
Next, the usable container space is partitioned along its height into stacked horizontal layers, each of which is further subdivided across the horizontal plane into non-overlapping regions that define the object target poses and collectively form a terminal packing layout.
A forward packing trajectory is then recovered by iteratively removing geometrically accessible objects from the layout and reversing the removal order, with the terminal object poses serving as placement targets.
The resulting packing trajectory is represented as
\begin{equation}
\mathcal{A}^*={(b_t^*,o_t^*,x_t^*,y_t^*)}_{t=0}^{T-1},
\end{equation}
where $\mathcal{A}^*$ denotes the resultant training packing trajectory, and $b_t^*,o_t^*,(x_t^*,y_t^*)$ denote the selected object, its orientation, and horizontal placement location, respectively.

Lastly, each reconstructed training packing trajectory is replayed and validated under gravity dynamics, rigid-body collision handling, and stability feedback, and packing trajectories containing invalid placements are discarded.
The scene states, placement actions, and physical feedback from valid rollouts are curated to form PackData-20K, a diverse collection of 20K training packing trajectories for closed-loop policy training.
PackData-20K spans the three difficulty levels described in Sec.~\ref{sec:test_organization}, and trajectories from different levels are mixed for robust policy training.

\subsection{\textit{PackLab-VLM}: Closed-Loop Policy Model}

\textit{PackLab-VLM} is a packing-specialized multimodal large
language model.
At each packing step, it conditions on the complete multi-turn interaction context, comprising the previous scene states and executed actions together with the current scene state, and predicts the packing action.

\vspace{+5pt}
\noindent \textbf{Multimodal State Encoding and Action Prediction.}
Let $\mathcal{S}_{k}=(\mathbf{H}_{k},\mathcal{B}_{k})$ denote the scene state at step $k$.
The model input is defined as a temporally ordered multimodal context:
\begin{equation}
\mathcal{X}_t=
\left(
\mathcal{P}_{\mathrm{sys}}, \mathcal{S}_1,\mathbf{a}_1,\ldots,
\mathcal{S}_{t-1},\mathbf{a}_{t-1},
\mathcal{S}_t
\right),
\label{eq:vlm_context}
\end{equation}
which includes a system prompt $\mathcal{P}_{\mathrm{sys}}$, historical heightmaps $\mathbf{H}_{<t}$, historical candidate-object buffers $\mathcal{B}_{<t}$, executed actions $\mathcal{A}_{<t}$, and the current observation $\mathcal{S}_{t}=(\mathbf{H}_t,\mathcal{B}_t)$.

The textual components of $\mathcal{X}_t$, including $\mathcal{B}_{\leq t}$ and $\mathcal{A}_{<t}$, are serialized in chronological order to form a structured prompt $Q_t$ with image placeholders and mapped to model token embeddings:
\begin{equation}
    \mathbf{U}_t = E_{\mathrm{text}}\!\left(\operatorname{Tok}(Q_t)\right) \in \mathbb{R}^{N_q \times d},
\label{eq:text_tokens}
\end{equation}
where $\operatorname{Tok}(\cdot)$ denotes the tokenizer, $E_{\mathrm{text}}(\cdot)$ denotes the token embedding layer, $N_q$ is the number of textual tokens, and $d$ is the embedding dimension.

Meanwhile, each current or historical container heightmap $\mathbf{H}_k$ is processed by a visual encoder $E_{\mathrm{visual}}(\cdot)$.
The resulting visual features are projected by a vision-language merger $M(\cdot)$ into the shared model embedding space:
\begin{equation}
    \mathbf{V}_k
    =
    M\!\left(E_{\mathrm{visual}}(\mathbf{H}_k)\right)
    \in \mathbb{R}^{N_v\times d},
    \label{eq:visual_tokens}
\end{equation}
where $N_v$ is the number of visual tokens and $d$ matches the model embedding dimension.
The visual tokens are inserted at the corresponding image-placeholder positions in the textual prompt, forming a shared multimodal token sequence:
\begin{equation}
    \mathbf{F}_t
    =
    \Phi
    \left(\mathbf{U}_t,\mathbf{V}_{\leq t}\right).
    \label{eq:multimodal_context}
\end{equation}
where $\mathbf{V}_{\leq t}=(\mathbf{V}_1,\ldots,\mathbf{V}_t)$ denotes the visual tokens from the current and historical steps, and $\Phi(\cdot)$ denotes placeholder-based multimodal token composition.

Finally, \textit{PackLab-VLM} autoregressively decodes the multimodal representation $\mathbf{F}_t$ into a structured textual output $\widehat{\boldsymbol{\tau}}_t$, which is parsed into an executable packing action $\mathbf{a}_t$:
\begin{equation}
    \begin{split}
    \widehat{\boldsymbol{\tau}}_t
    & =
    \Psi(\mathbf{F}_t), \\
    \qquad
    \mathbf{a}_t
    &  =
    (b_t,o_t,x_t,y_t)
    =
    \operatorname{Parse}
    \left(\widehat{\boldsymbol{\tau}}_t\right).
    \end{split}
    \label{eq:action_parsing}
\end{equation}
where $\Psi(\cdot)$ denotes the standard token-level autoregressive decoding process of \textit{PackLab-VLM}.

\vspace{+5pt}
\noindent \textbf{Policy Training.} \
\textit{PackLab-VLM} is fine-tuned on step-level multimodal examples extracted from the training packing trajectories in the PackData-20K dataset, denoted by
\begin{equation}
    \mathcal{D}
    =
    \left\{
    \left(\mathcal{X}_{n,t},\boldsymbol{\tau}_{n,t}^{*}\right)
    \mid
    1\leq n\leq N,\ 1\leq t\leq T_n
    \right\},
    \label{eq:sft_dataset}
\end{equation}
where $n$ is the index of a training packing trajectory, $t$ is the packing-step index, $\mathcal{X}_{n,t}$ denotes the multimodal state--history context at step $t$, and $\boldsymbol{\tau}_{n,t}^{*}$ denotes the corresponding ground-truth textual action output formed by the training packing trajectory $\mathcal{A}^*$.
Each trajectory provides supervision at its decision steps, training the MLLM to generate executable structured action text.

The supervised fine-tuning objective is the target-token negative log-likelihood:
\begin{equation}
    \mathcal{L}
    =
    -\frac{1}{N}
    \sum_{n=1}^{N}
    \sum_{t=1}^{T_n}
    \log
    p_{\Theta}
    \left(
        \boldsymbol{\tau}_{n,t}^{*}
        \mid
        \mathcal{X}_{n,t}
    \right),
    \label{eq:sft_loss}
\end{equation}
where $T_n$ denotes the total number of packing steps in the $n$-th training trajectory, $\Theta$ denotes the trainable parameters of \textit{PackLab-VLM}, and $p_{\Theta}(\cdot)$ is the token probability distribution produced by the model.
The probability is factorized autoregressively over the target action-output tokens.

\begin{table}
\caption{Statistics of \textit{PackLab-Bench} test cases by
difficulty.}
\label{tab:evaluation_cases}
\centering
\setlength{\tabcolsep}{5mm}
\renewcommand{\arraystretch}{1.2}
\begin{tabular}{lccc}
\specialrule{0.1em}{0pt}{2pt}
Difficulty & Buffer & Layers & Objects \\
\midrule
Easy
& 3 & 1--2 & 4--8 \\
Medium
& 5 & 2--3 & 9--18 \\
Hard
& 10 & 3--4 & 19--32 \\
\specialrule{0.1em}{1pt}{0pt}
\end{tabular}
\end{table}

\begin{table*}[t]
    \caption{Quantitative results on \textit{PackLab-Bench}. We compare \textit{PackLab-VLM} with representative packing methods across easy, medium, hard, and average settings. Each setting reports Success Ratio (Succ.), Compactness (Comp.), and Overall Score (Overall). The best average results are marked in \textbf{bold}.}
    \label{tab:main_results}
    \centering
    \setlength{\tabcolsep}{2.0mm}
    \renewcommand{\arraystretch}{1.15}
    \resizebox{\linewidth}{!}{
    \begin{tabular}{l|ccccccccc|ccc}
        \specialrule{0.1em}{0pt}{2pt}
        \multicolumn{1}{l}{\multirow{2}{*}{Method}} &
        \multicolumn{3}{c}{Easy} &
        \multicolumn{3}{c}{Medium} &
        \multicolumn{3}{c}{Hard} &
        \multicolumn{3}{c}{Average} \\
        \cmidrule(lr){2-4}
        \cmidrule(lr){5-7}
        \cmidrule(lr){8-10}
        \cmidrule(lr){11-13}
        \multicolumn{1}{l}{} & Succ.$\uparrow$ & Comp.$\uparrow$ & Overall$\uparrow$
        & Succ.$\uparrow$ & Comp.$\uparrow$ & Overall$\uparrow$
        & Succ.$\uparrow$ & Comp.$\uparrow$ & \multicolumn{1}{c}{Overall$\uparrow$}
        & Succ.$\uparrow$ & Comp.$\uparrow$ & Overall$\uparrow$ \\
        \midrule
        DBLF~\cite{karabulut2004hybrid} & 0.788 & 0.690 & 0.568 & 0.967 & 0.645 & 0.626 & 0.720 & 0.641 & 0.469 & 0.825 & 0.659 & 0.554 \\
        HM~\cite{wang2019stable,wang2021dense} & 0.788 & 0.690 & 0.568 & 0.967 & 0.639 & 0.619 & 0.680 & 0.601 & 0.415 & 0.812 & 0.643 & 0.534 \\
        MTA~\cite{wang2010two} & 0.764 & 0.654 & 0.543 & 0.895 & 0.540 & 0.489 & 0.571 & 0.487 & 0.285 & 0.743 & 0.560 & 0.439 \\
        FFP~\cite{berkey1987two} & 0.771 & 0.673 & 0.558 & 0.698 & 0.389 & 0.278 & 0.440 & 0.362 & 0.162 & 0.636 & 0.475 & 0.333 \\
        SDF-Pack~\cite{pan2023sdf} & 0.880 & 0.795 & 0.718 & 0.966 & 0.640 & 0.619 & 0.719 & 0.646 & 0.469 & 0.855 & 0.694 & 0.602 \\
        TAP-Net~\cite{hu2020tap} & 0.890 & 0.555 & 0.501 & 0.972 & 0.548 & 0.535 & 0.673 & 0.566 & 0.386 & 0.845 & 0.556 & 0.474 \\
        IR-BPP~\cite{zhao2021online} & 0.695 & 0.592 & 0.434 & 0.916 & 0.565 & 0.520 & 0.656 & 0.559 & 0.372 & 0.756 & 0.572 & 0.442 \\
        \rowcolor{ourslightblue} \textit{PackLab-VLM} (Ours) & 0.999 & 0.923 & 0.922 & 0.966 & 0.680 & 0.665 & 0.680 & 0.552 & 0.394 & \textbf{0.882} & \textbf{0.718} & \textbf{0.660} \\
        \specialrule{0.1em}{1pt}{0pt}
    \end{tabular}
    }
\end{table*}

\begin{figure*}
    \centering
    \includegraphics[width=\linewidth]{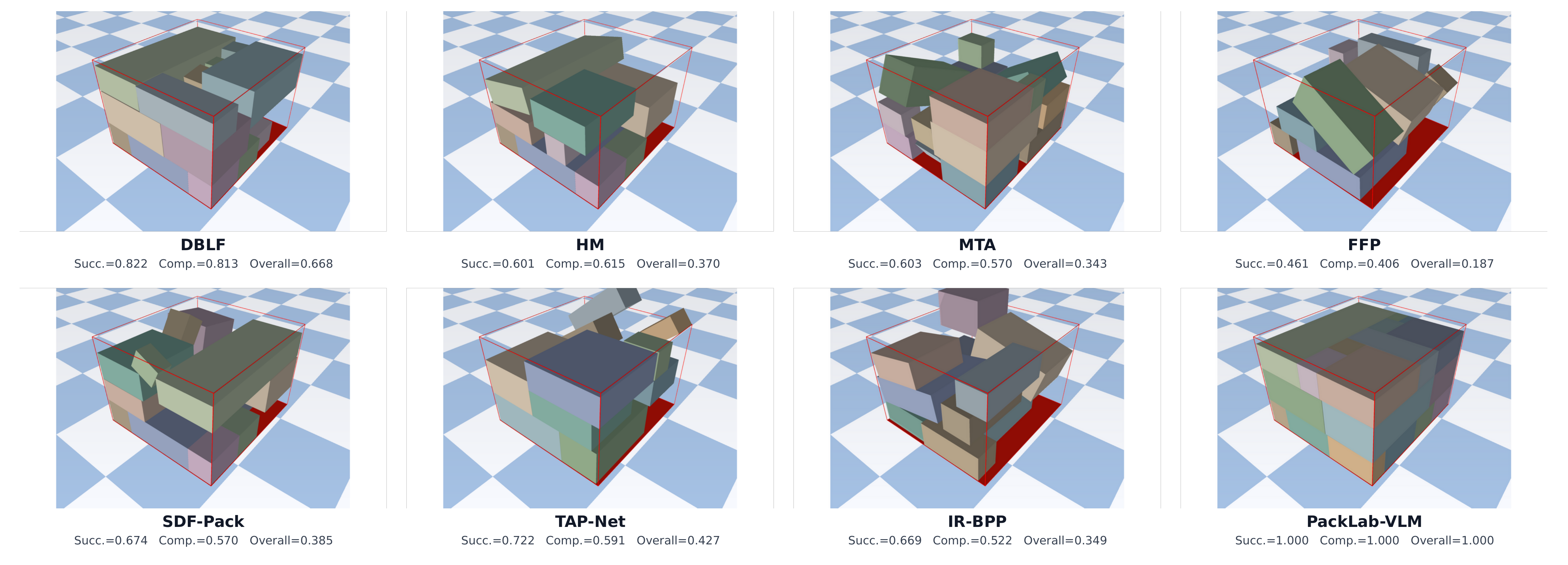}
    \caption{Qualitative comparison of \textit{PackLab-VLM} with existing methods on an example packing case. \textit{PackLab-VLM} produces a complete and compact arrangement with all objects contained within the container, achieving 1.000 on all three metrics, whereas competing methods yield less compact layouts with lower Success Ratio.}
    \label{fig:virtual}
\end{figure*}

\subsection{\textit{PackLab-Bench}: Standardized Evaluation Protocol}

\vspace{+5pt}
\noindent \textbf{Test Cases Organized by Difficulty Levels.}
\label{sec:test_organization}
\textit{PackLab-Bench} contains 60 test cases, evenly
divided among easy, medium, and hard levels, spanning 905 packing steps in total.
As summarized in Table~\ref{tab:evaluation_cases}, task difficulty
is defined by jointly varying the candidate-buffer size,
number of packing layers and objects, and container dimensions,
providing controlled coverage of different decision horizons and
spatial configurations.
\textit{PackLab-Bench} is strictly disjoint from PackData-20K, as all test cases are independently sampled and carefully verified to exclude any overlap.

\vspace{+5pt}
\noindent \textbf{Evaluation Metrics.}
We evaluate each method with three metrics after the completion of each packing case: Success Ratio, Compactness, and Overall Score.
Let $\mathcal{O}$ denote the complete object set and $\mathcal{O}_{\mathrm{succ}}$ denote the subset of successfully packed objects whose post-placement geometric centers lie in the container boundaries.
The Success Ratio is
\begin{equation}
    R_{\mathrm{succ}} =
    \frac{\sum_{b \in \mathcal{O}_{\mathrm{succ}}} V_b}
    {\sum_{b \in \mathcal{O}} V_b},
\end{equation}
where $V_b$ denotes the volume of a packed object $b$.
Compactness is measured by the ratio of the total volume of successfully packed objects to the packing-envelope volume, defined by the container base area and the maximum occupied height in the final heightmap:
\begin{equation}
    C =
    \frac{\sum_{b \in \mathcal{O}_{\mathrm{succ}}} V_b}
    {W \cdot L \cdot \max_{x,y} H_{T+1}[x,y]},
\end{equation}
where $W$ and $L$ are the container's width and length, and $H_{T+1}$ is the terminal heightmap.
Overall Score is the product of Success Ratio and Compactness, defined as
\begin{equation}
    R_{\mathrm{all}} = R_{\mathrm{succ}} \cdot C.
\end{equation}
It serves as the primary metric for jointly evaluating how much volume is packed and how compactly it is arranged.

\section{Experiments}
\label{sec:experiments}

\begin{figure}[t]
    \centering
    \includegraphics[width=\linewidth, height=0.6\linewidth]{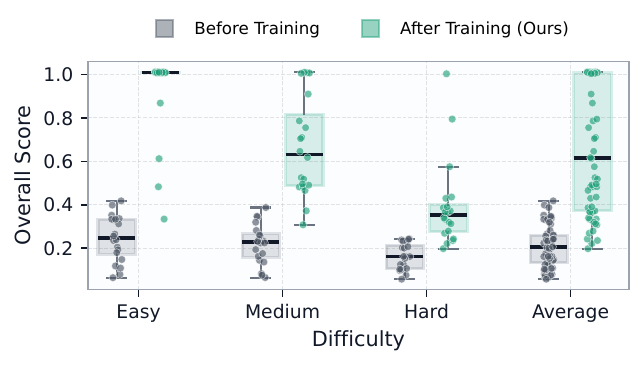}
    \caption{Distributions of packing performance before and after training across different difficulty levels.
    Each scatter point denotes the Overall Score obtained for an individual packing case, while the box plots report the median and interquartile range for each difficulty level.
    The post-training scores are consistently higher than the corresponding pre-training scores across all difficulty levels, indicating improved packing performance after task-specific training.}
    \label{fig:base_sft_score_boxplot}
\end{figure}

\subsection{Experimental Setup}
\noindent\textbf{Implementation Details.}
We instantiate \textit{PackLab-VLM} with Qwen3.5-9B~\cite{qwen3.5-9b} as the backbone and train it on PackData-20K using packing-oriented supervised fine-tuning.
We discretize $o_t$ as 0 and 1 to represent $0^\circ$ and $90^\circ$, respectively, and discretize $(x_t,y_t)$ at $2.5\ \mathrm{cm}$ resolution.
Optimization uses AdamW with a fixed learning rate of $1\times10^{-5}$ and a global batch size of 8, where each GPU processes one training sample per step.
We train the model for a total of 5 epochs on an 8-GPU training cluster.
To make long-context multimodal training feasible in practice, we use Fully Sharded Data Parallel (FSDP), \texttt{bf16} precision, and gradient checkpointing during training.

\vspace{+5pt}
\noindent\textbf{Compared Methods.}
We compare with five classic placement heuristics,
as well as two reinforcement learning (RL)-based methods using their official checkpoints.
Deepest-Bottom-Left-Fill (DBLF)~\cite{karabulut2004hybrid} favors low-corner placements for compact bottom-up stacking.
Heightmap-Minimization heuristic (HM)~\cite{wang2019stable,wang2021dense} evaluates candidate placements by heightmap changes to maintain a low, smooth surface.
Maximum Touching Area (MTA)~\cite{wang2010two} maximizes contact area with the floor, walls, or packed objects.
First Feasible Placement (FFP)~\cite{berkey1987two} returns the first feasible position in the search order.
SDF-Pack~\cite{pan2023sdf} minimizes signed distance field values to favor spatially compact placements.
TAP-Net~\cite{hu2020tap} learns a transport-and-pack policy for sequential packing, while IR-BPP~\cite{zhao2021online} learns an online packing policy with a reinforcement-learning objective.
For these two learning-based baselines, we evaluate the official checkpoints with the required action-interface adaptation.

\subsection{Main Results}
Table~\ref{tab:main_results} reports the quantitative results on \textit{PackLab-Bench}.
All methods are evaluated on the same fixed test cases, and the reported results are averaged over three independent runs
with different random seeds.
Overall, \textit{PackLab-VLM} achieves the best average performance among all compared methods, with an average Overall Score of 0.660, outperforming the strongest heuristic baseline, SDF-Pack, by 0.059, and the strongest official RL checkpoint, TAP-Net, by 0.186.
It also obtains the highest average Success Ratio and Compactness scores, reaching 0.882 and 0.718, respectively, both higher than SDF-Pack, TAP-Net, and IR-BPP.
This indicates that the improvement is not driven by a single metric but by jointly placing more object volume inside the container and arranging it more compactly.

Across difficulty levels, \textit{PackLab-VLM} performs particularly strongly on easy and medium tasks.
On easy cases, it achieves nearly complete Success Ratio and a substantially higher Overall Score than all baselines, showing that the model can effectively leverage visual state, object information, and packing history when spatial constraints are moderate.
On medium cases, its Success Ratio is comparable to the best heuristic and RL baselines, while its higher Compactness leads to the best Overall Score, suggesting better spatial organization rather than merely placing more objects.
Hard cases remain more challenging due to tighter space constraints and more complex residual layouts.
\textit{PackLab-VLM} still achieves competitive performance in this setting, while geometry-search heuristics such as SDF-Pack and DBLF retain advantages in some densely constrained cases.

These results demonstrate the effectiveness of \textit{PackLab-VLM} over greedy packing heuristics and RL policies optimized for fixed configurations.
For a controlled comparison, we retrain TAP-Net~\cite{hu2020tap} and IR-BPP~\cite{zhao2021online} under the same task distribution, using an equivalent number of trial-and-error iterations.
Across three test-time seeds, TAP-Net and IR-BPP achieve average Overall Scores of 0.405 and 0.345, respectively, underscoring the difficulty of learning a unified RL policy across heterogeneous packing configurations.

\begin{table}[t]
    \caption{Ablation results on input context.}
    \label{tab:ablation_input_context}
    \centering
    \setlength{\tabcolsep}{2.8mm}
    \renewcommand{\arraystretch}{1.12}
    \begin{tabular}{l|lc}
        \specialrule{0.1em}{0pt}{2pt}
        \multicolumn{1}{l}{Aspect} & Variant & Overall Score $\uparrow$ \\
        \midrule
        \multirow{2}{*}{Visual Input} & Distractor Heightmap & 0.494 \\
        & \cellcolor{ourslightblue}Original Heightmap (Ours) & \cellcolor{ourslightblue}\textbf{0.660} \\
        \midrule
        \multirow{3}{*}{Sequential History} & Only Current Observation & 0.533 \\
        & W/o Action History & 0.589 \\
        & \cellcolor{ourslightblue}Full History (Ours) & \cellcolor{ourslightblue}\textbf{0.660} \\
        \specialrule{0.1em}{1pt}{0pt}
    \end{tabular}
\end{table}

In Figure~\ref{fig:virtual}, we also provide qualitative comparisons among all methods.
The visualized packing trajectories and final configurations show that \textit{PackLab-VLM} generally produces more coherent arrangements, makes better use of the available space, and preserves usable regions for subsequent objects.
In contrast, heuristic methods often favor locally feasible placements that can fragment the remaining space and limit future packing opportunities.
Although RL-based methods can outperform some heuristics in certain cases, their layouts are still less complete and compact than \textit{PackLab-VLM} under heterogeneous packing configurations.
These examples illustrate how incorporating visual observations, object attributes, and packing history enables \textit{PackLab-VLM} to reason about both immediate placement feasibility and the longer-term organization of the container.
More results are presented in the supplementary video.

\begin{table}[t]
    \caption{Ablation results on training data.}
    \label{tab:ablation_data}
    \centering
    \setlength{\tabcolsep}{2.8mm}
    \renewcommand{\arraystretch}{1.12}
    \begin{tabular}{l|lc}
        \specialrule{0.1em}{0pt}{2pt}
        \multicolumn{1}{l}{Aspect} & Variant & Overall Score $\uparrow$ \\
        \midrule
        \multirow{3}{*}{Data Scale} & 25\% Training Data & 0.578 \\
        & 50\% Training Data & 0.630 \\
        & \cellcolor{ourslightblue}100\% Training Data (Ours) & \cellcolor{ourslightblue}\textbf{0.660} \\
        \midrule
        \multirow{3}{*}{Data Recipe} & Easy-Only Recipe & 0.587 \\
        & Easy+Medium Recipe & 0.604 \\
        & \cellcolor{ourslightblue}Mixed Recipe (Ours) & \cellcolor{ourslightblue}\textbf{0.660} \\
        \specialrule{0.1em}{1pt}{0pt}
    \end{tabular}
\end{table}

\begin{figure*}
    \centering
    \includegraphics[width=\linewidth]{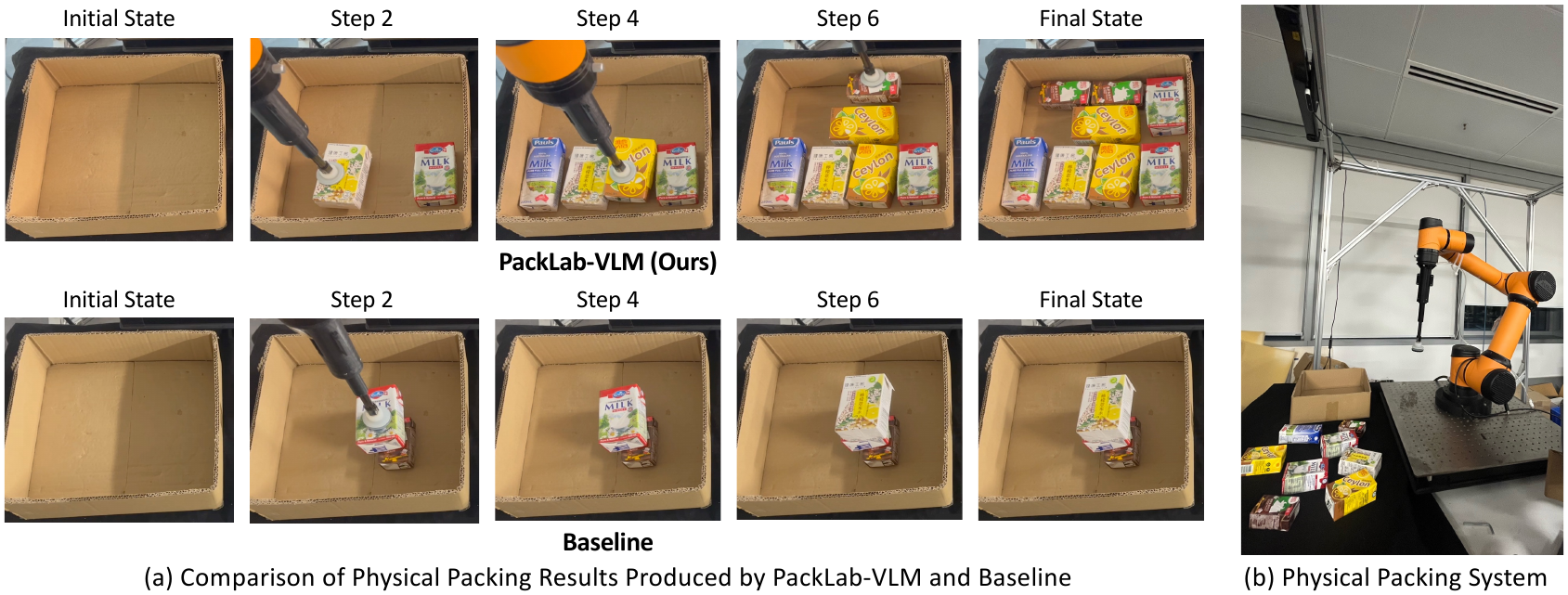}
    \caption{Qualitative comparison on the same real-world packing case of eight objects.
    (a) Sequential snapshots show that \textit{PackLab-VLM} (top) successfully packs all objects into the container with a compact, space-efficient arrangement, whereas the baseline (bottom) packs only a subset of three objects and leaves a large portion of the container unused.
    (b) The physical robotic packing system used for the real-world experiment.}
    \label{fig:demo}
\end{figure*}

\subsection{Ablation Studies and More Analysis}
We perform ablation studies on \textit{PackLab-VLM}'s training and input context, along with additional analysis on the training data and model size.
Following the main experiments, all results are averaged over three runs.

\vspace{+5pt}
\noindent\textbf{Training Effect.}
We first examine whether the gain of \textit{PackLab-VLM} comes from packing-oriented training rather than the zero-shot capability of the Qwen3.5-9B backbone.
We directly compare the base model and the trained model under the same benchmark setting.
As shown in Figure~\ref{fig:base_sft_score_boxplot}, training substantially improves the average Overall Score from 0.209 to 0.660.
The improvement is consistent across difficulty levels: Easy increases from 0.246 to 0.922, Medium from 0.220 to 0.665, and Hard from 0.162 to 0.394.
The cross-case distributions also shift upward after training, indicating that the gain is not caused by a few favorable cases.
These results confirm that task-specific supervision is essential for learning valid action formats, object selection, and spatially grounded placement behavior.

\vspace{+5pt}
\noindent\textbf{Input Context.}
We study whether \textit{PackLab-VLM} relies on input context rather than superficial input formatting.
For visual input, we keep the trained model and inference pipeline unchanged, but replace the original heightmap with a black distractor heightmap at inference time.
For sequential history, we compare full history with two reduced settings: using only the current observation and removing assistant action history.
As shown in Table~\ref{tab:ablation_input_context}, replacing the original heightmap reduces Overall Score from 0.660 to 0.494, confirming that visual geometry provides essential spatial cues.
Removing sequential history also hurts performance: Only Current Observation scores 0.533, while W/o Action History reaches 0.589, both below the full-history setting.
These results show that \textit{PackLab-VLM} benefits from both the visual and sequential history input designs, supporting the effectiveness of our multimodal closed-loop formulation.

\vspace{+5pt}
\noindent\textbf{Training Data.}
We next explore whether \textit{PackLab-VLM} benefits from more training data and a broader difficulty recipe.
For data scale, we keep the architecture and evaluation protocol fixed while using 25\%, 50\%, and 100\% of the training data.
Overall Score improves steadily from 0.578 to 0.630 and 0.660, showing that additional training packing trajectories provide useful coverage beyond basic action-format learning.
For the data recipe, Easy-Only reaches 0.587, while Easy+Medium improves to 0.604 by adding more constrained states.
However, both are below the mixed easy/medium/hard recipe, which achieves 0.660.
This indicates that diverse difficulty coverage is important for learning robust packing behavior: easy cases stabilize basic geometric grounding, medium cases teach common constrained decisions, and hard cases expose dense, high-ambiguity states.
Together, these results support the effectiveness of our full training recipe.

\begin{figure}
    \centering
    \includegraphics[width=\linewidth]{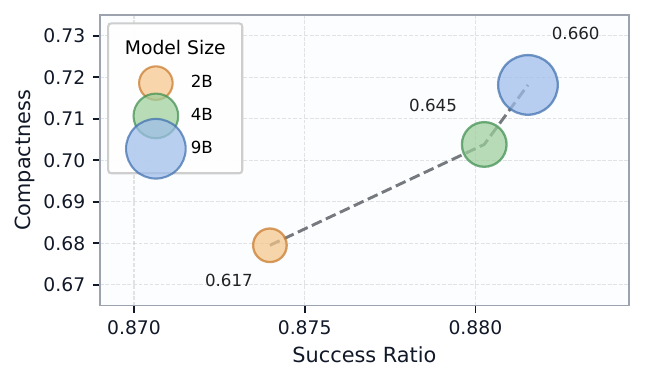}
    \caption{Model-size ablation of trained models. Each point reports Success Ratio and Compactness, with nearby text showing the Overall Score.}
    \label{fig:model_size_ablation}
\end{figure}

\vspace{+5pt}
\noindent\textbf{Model Size.}
Finally, we examine how model capacity affects the trained packing policy.
Figure~\ref{fig:model_size_ablation} compares trained 2B, 4B, and 9B models using Success Ratio and Compactness as axes, with Overall Score annotated near each point.
Due to computational resource constraints, we use 9B as the largest model in this study.
All three models achieve a strong Success Ratio after task-specific training, while larger models consistently improve Compactness and Overall Score.
Specifically, the Overall Score increases from 0.617 for 2B to 0.645 for 4B and 0.660 for 9B.
This trend suggests that larger models better integrate heightmaps, object attributes, and sequential history when organizing compact placements, rather than merely increasing the amount of packed volume.
Larger MLLMs remain an important direction for future work.

\subsection{Evaluation on a Physical Platform}
To assess the real-world applicability of \textit{PackLab-VLM}, we developed a physical platform that performs scene perception, packing planning, and robotic execution, with re-planning triggered when the observed container state deviates substantially from the prediction.
The platform comprises an AUBO robot arm with a suction cup, a RealSense RGB camera, a Photoneo depth camera, a buffer area with randomly posed objects, and a $30\ \mathrm{cm}\times30\ \mathrm{cm}\times10\ \mathrm{cm}$ container.

We use SAM3~\cite{carion2026sam} to detect and segment all buffered objects, whose point clouds are extracted to estimate their 3D dimensions and initialize the virtual environment.
\textit{PackLab-VLM} then selects objects and predicts their target poses.
For each object, the segmented point cloud provides a top-center grasp and 2D orientation through principal-axis analysis.
The robot aligns this axis with the container $y$-axis and executes the placement.
After each placement, the container is re-scanned to determine whether re-planning is required.

Figure~\ref{fig:demo} shows the physical setup and compares \textit{PackLab-VLM} with the baseline using the same object set.
The results show that \textit{PackLab-VLM} produces a more compact packing arrangement than the baseline, resulting in more effective utilization of the container space.
Additional results are provided in the supplementary video.

\section{Conclusion}

In this paper, we study MLLM-based robotic bin packing under a multimodal decision-making paradigm.
To address this problem, we propose \textbf{PackLab}, a comprehensive framework including \textit{PackLab-Suite} for physics-based simulation and data generation, \textit{PackLab-VLM} for closed-loop object selection and placement prediction, and \textit{PackLab-Bench} for standardized evaluation across tasks of varying difficulty.
Extensive experiments show that \textit{PackLab-VLM} outperforms conventional packing heuristics, reinforcement learning methods, and general-purpose MLLMs, demonstrating the effectiveness of packing-specific training and evaluation for improving MLLM-based robotic bin packing.
Future work will extend PackLab to broader object geometries and physical settings, and enable more flexible packing behaviors through user-defined objectives and constraints.

\bibliographystyle{IEEEtran}
\bibliography{egbib}

\end{document}